\documentclass[letterpaper]{article}
\usepackage{aaai19}  
\usepackage{times}  
\usepackage{helvet}  
\usepackage{courier}  
\usepackage[hyphens]{url} 
\usepackage{graphicx}  
\usepackage{placeins}

\usepackage{colortbl}
\usepackage{nth}
\usepackage{capt-of}
\usepackage{comment}
\usepackage{multirow}
\usepackage{floatrow}
\usepackage{float}
\usepackage{booktabs}
\usepackage{enumitem}
\usepackage{multicol}
\usepackage{graphicx}
\usepackage{subcaption}
\usepackage{amsmath}

\usepackage{arydshln}

\begin{document}
\title{Do All Good Actors Look The Same? Exploring News Veracity Detection Across The U.S. and The U.K.}


\author{Benjamin D. Horne, Maur\'{i}cio Gruppi, and Sibel Adal{\i}\\
 Rensselaer Polytechnic Institute\\
 \{horneb,gouvem,adalis\}@rpi.edu}

\maketitle

\begin{abstract} 
A major concern with text-based news veracity detection methods is that they may not generalize across countries and cultures. In this short paper, we explicitly test news veracity models across news data from the United States and the United Kingdom, demonstrating there is reason for concern of generalizabilty. Through a series of testing scenarios, we show that text-based classifiers perform poorly when trained on one country's news data and tested on another. Furthermore, these same models have trouble classifying unseen, unreliable news sources. In conclusion, we discuss implications of these results and avenues for future work. 
\end{abstract}

\section{Introduction}
A concern with text-based news veracity classifiers is that they may not generalize across real-life situations. In Machine Learning, guarantees are based on the assumption that the data used for training a model comes from the same distribution as the data for the model's eventual use (test data). In news veracity detection, this means we are relying on the higher level labeling concepts, such as reliability, bias, true, and false, to manifest consistently across the data points. This consistency across data point is of course dependent on the features or representation used. However, with news production (and many other socially influenced processes) this assumption may not hold and can be difficult to assess. Hence, we need to design experiments to explicitly test when and how predictions hold across different real life settings. 

To a certain extent these types of experiments have been done. For example, recent work has tested news veracity classifiers performance over time, demonstrating that text-based classifiers do degrade in performance over time~\cite{horne2019robust}. The same study also experimented with news veracity classifiers under various malicious content manipulation attacks. However, there are many other real life settings that have not been studied. One real life setting that has not been explicitly tested is generalization across different countries news. While there has been a considerable amount of work on news veracity detection, the focus has been on news from the United States or the country of each news source has not been taken into account. Hence, in this work we ask the question:

\textbf{Q:} Do text-based news veracity models trained on one countries news data generalize to another?

Specifically, in this short paper we use the United States (US) and the United Kingdom (UK) as our case study. We choose this test-bed due to both the availability of data in both countries and the use English in both, which allows us to test state-of-the-art news veracity models across both. 

Our results indicate that text-based models trained on US news data perform poorly when classifying UK news data, and vice-versa. Our results also show that text-based models can perform poorly on unseen, unreliable sources, pointing to a need for ample unreliable data and for new features based on signals outside of the news article text.

\section{Related Work}
There are very few works that explore news veracity detection across different countries. A 2018 study by Gruppi et al. explores text features across reliable and unreliable news sources in both the U.S. and Brazil~\cite{gruppi2018exploration}. The main goal was to understand how text features commonly used in news veracity detection change across U.S. English and Brazilian Portuguese. The authors show that a small set of writing style features work well across both languages; however, not all types of features work across both. In this paper, we focus on two countries that use the same language in news reports (i.e. the United States and the United Kingdom). Because of this single language use, we are able to explore a more complete, state-of-the-art set of text features in our analysis.

There are more general works that focus on changes in news coverage between different countries, typically over specific events~\cite{wu2000systemic}. Many of the studies in this area show significant differences between news in different countries, indicating that text-based news veracity models may perform poorly across countries. 

\section{Data}
To explore this problem, we extract news article data from the NELA-GT-2018 dataset~\cite{norregaard2019nela}. We extract three classes of news sources: reliable sources from the United States (\texttt{US}), reliable sources from the United Kingdom (\texttt{UK}), and unreliable sources, regardless of location (\texttt{UR}). While the NELA-GT-2018 dataset has many veracity labels, the U.K. is not very well covered. Due to this, we group sources in the following way: 
\begin{enumerate}
    \item Using Media Bias/Fact Check (MBFC) factuality scores in the NELA-GT-2018 dataset, we select all sources headquartered in the United Kingdom that have a factuality score of four or five, where five is the max factuality score. 
    \item Again using the MBFC factuality, we randomly select sources headquartered in the United States with factuality scores of four or five equal to the number of sources selected from the United Kingdom.
    \item Lastly, we randomly select sources that have a MBFC factuality score of one or two, where one is the lowest score, equal to the number of sources selected from the United Kingdom, regardless of where the sources are located (often the location of these low quality sources is not known).
\end{enumerate}

This process gives us five sources in each of the three classes, which can be found in Table~\ref{tab:sources}. While the total number of sources is much less than the number of sources used in other news veracity studies, our goal is to have a balanced number of sources in each class. 

\begin{table}[]
    \centering
    \begin{tabular}{c|c|c}
       \textbf{US}  &  \textbf{UK} & \textbf{UR}\\
       \hline 
        ABC News & The Telegraph & Infowars\\
        NPR & The Independent & Waking Times\\
        USA Today & The Guardian UK & Intellihub\\
        CBS News & BBC UK & True Pundit\\
        PBS & The Evening Standard & Activist Post\\
    \end{tabular}
    \caption{US - Reliable sources from the United States, UK - Reliable Sources from the United Kingdom, UR - Unreliable sources, regardless of location. }
    \label{tab:sources}
\end{table}

\section{Feature Models}
To build our text-based news veracity models, we use two types of feature sets, one hand-crafted feature set commonly used in news veracity studies and one automatic feature extraction method. Specifically, we use:
\begin{itemize}
    \item NELA - NELA is a set of hand-crafted features that can be categorized into five categories: style, complexity, bias, affect, and moral. In total the feature set contains 194 features. The feature set was introduced in \cite{horne2018accessing} and has been used in multiple news veracity studies since \cite{baly2018,cruz2019team}.
    \item Doc2Vec - Another way to capture differences in text is to use an automatic feature extraction technique. In this case we use a paragraph embedding model, Doc2Vec~\cite{le2014distributed} with 100 dimension representation. 
\end{itemize}

We also test the scaled and normalized versions of these feature sets, called NELA-scaled and Doc2Vec-scaled.

\section{Model Testing}~\label{sec:testing}
With these feature sets, we train and test classification models using Random Forest and three different testing schemes:
\begin{itemize}
    \item Article Split - In this setting we remove 20\% of the articles from the dataset for testing, a traditional text classification setting of 80\% data for training and 20\% of data for testing. We do this over 20 folds and take the average. In the context of our news veracity classifier, this setting is simulating a model in which incoming news sources have been seen by the model (an ideal situation for the classifier). This can also be thought of as weak labeling.
    \item Source Split - In this setting remove all articles from 20\% of the sources for testing. We do this over 20 folds and take the average. In the context of our news veracity classifier, this setting is simulating a model in which incoming news articles are from news sources the model has never seen (a less ideal, but more realistic situation).
    \item Country Split - In this setting we train the classifier on all data from one country in the positive class (models can be trained as US versus UR or UK versus UR) and test the classifier on all the data from the other country (test on US or UK data). This setting explicitly tests the situation where a classifier was built on one country's news data and is deployed on another country's news data. 
\end{itemize}

In all testing schemes, we uniformly at random sample 1000 articles from each source, in each class to ensure balance. In practice, the balance between the sources has little impact on the article split and country split scenarios, but has impact on the source split scenario, as a different number of articles in each fold can inflate the measured accuracy.

In addition to these testing schemes, we test several different machine learning algorithms, including Random Forest, Extra Trees, and Support Vector Machine (SVM). As mentioned, we test both when the feature vectors are scaled and normalized and when they are not. Importantly, we only test the unscaled features for the decision tree based models (Random Forest and Extra Trees). We do this because decision trees can handle unscaled features, while SVMs cannot. In addition, past work in news veracity detection has shown high performance using Random Forest with unscaled, hand-crafted features~\cite{horne2019robust}. Due to the high similarity between the Random Forest and Extra Tree results, we only show the Random Forest.


\begin{figure}[h]
    \centering
   \begin{subfigure}{.49\linewidth} 
        \label{ref_label1}
        \centering
        \includegraphics[scale=0.2]{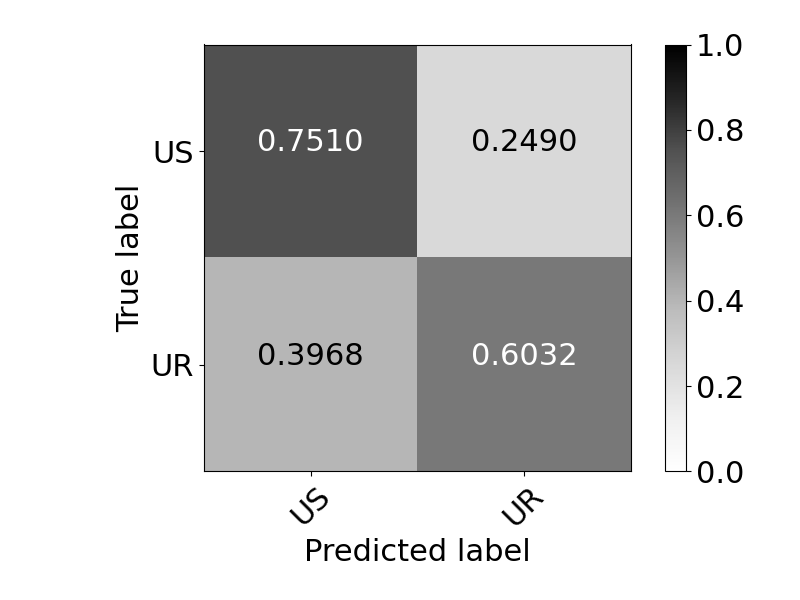}
        \caption{US vs. UR}
    \end{subfigure}
    \begin{subfigure}{.49\linewidth}
        \label{ref_label3}
        \centering
        \includegraphics[scale=0.2]{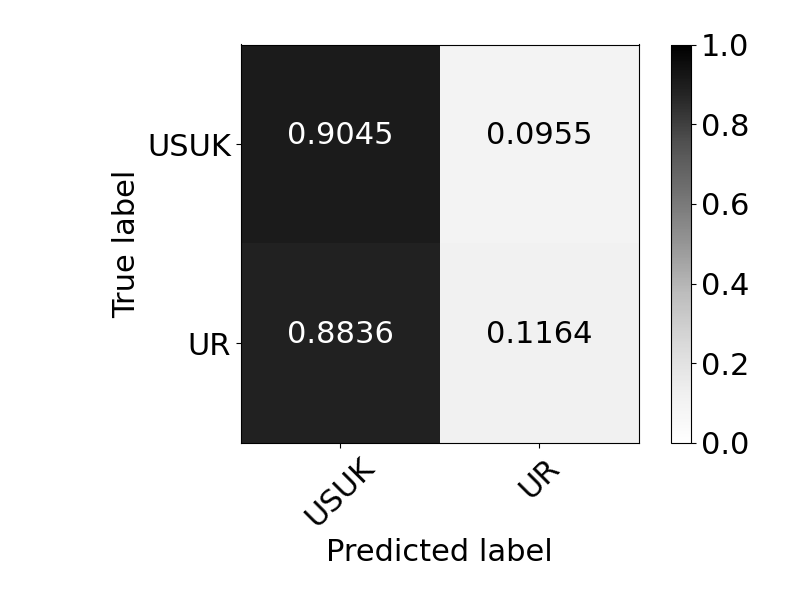}
        \caption{US+UK vs. UR}
    \end{subfigure}\\
    \begin{subfigure}{.49\linewidth}
        \label{ref_label1}
        \centering
        \includegraphics[scale=0.2]{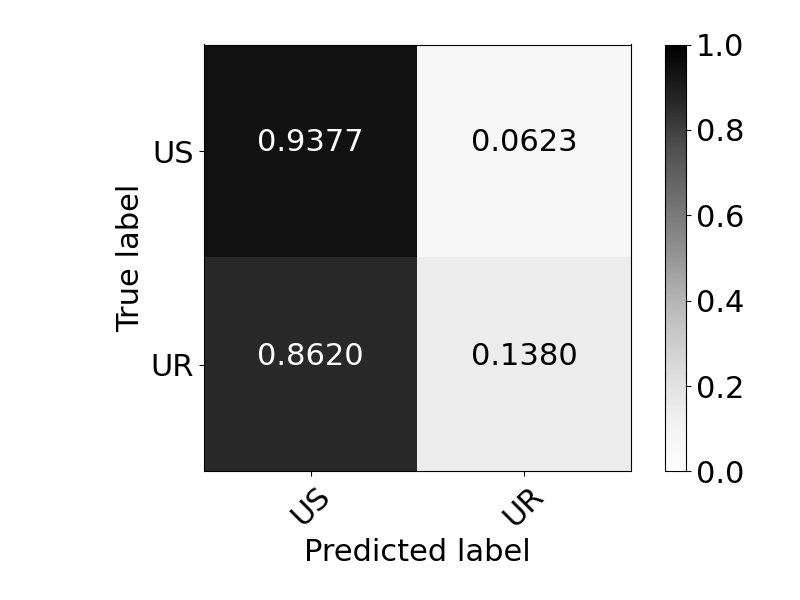}
        \caption{US vs. UR}
    \end{subfigure}
    \begin{subfigure}{.49\linewidth}
        \label{ref_label3}
        \centering
        \includegraphics[scale=0.2]{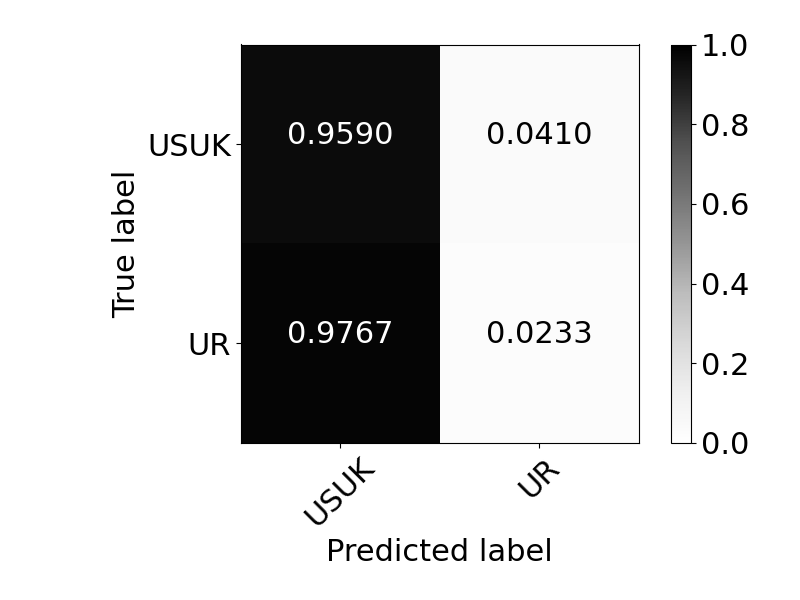}
        \caption{US+UK vs. UR}
    \end{subfigure}\\
    \begin{subfigure}{.49\linewidth}
        \label{ref_label1}
        \centering
        \includegraphics[scale=0.2]{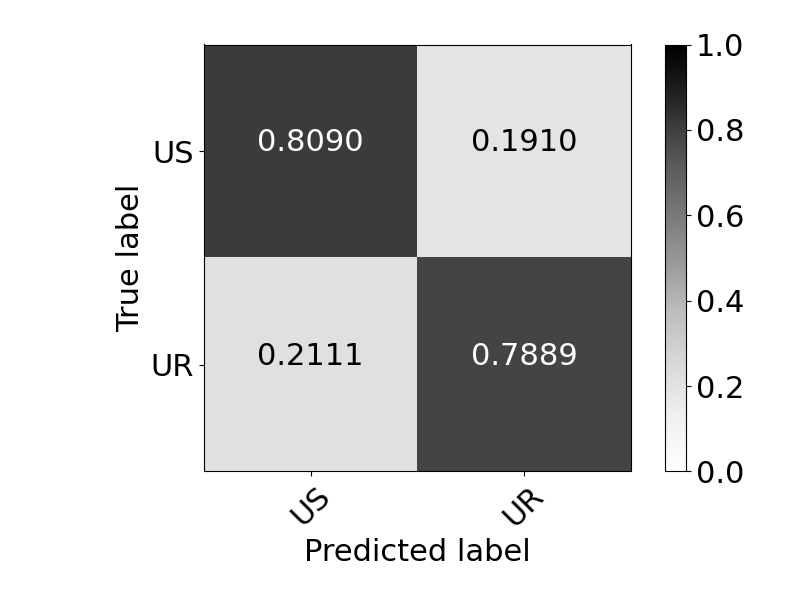}
        \caption{US vs. UR}
    \end{subfigure}
    \begin{subfigure}{.49\linewidth}
        \label{ref_label3}
        \centering
        \includegraphics[scale=0.2]{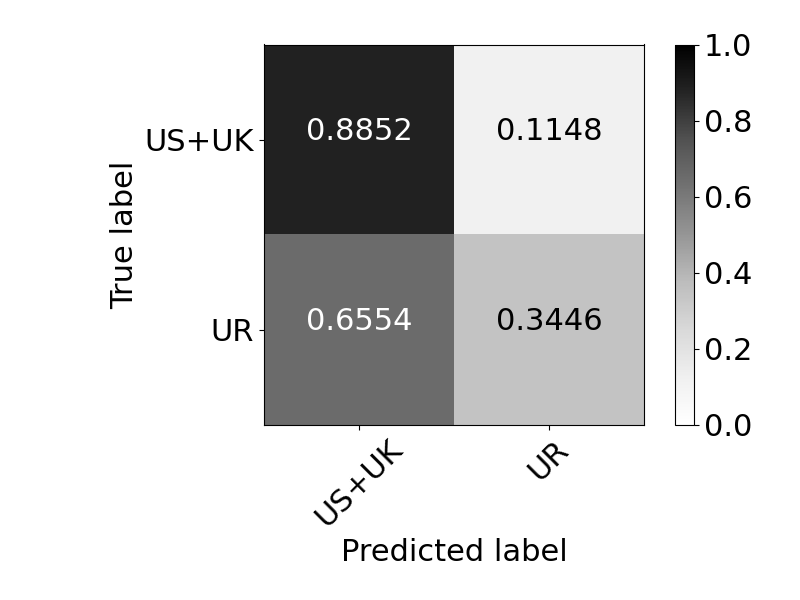}
        \caption{US+UK vs. UR}
    \end{subfigure}
    \caption{The average, normalized confusion matrix over each fold in the source split scenario. In row 1 (a and b), the model is Random Forest using NELA features. In row 2 (c and d), the model is Random Forest using NELA-scaled features. In row 3 (e and f), the model is SVM using NELA-scaled features.}
    \label{fig:conf}
\end{figure}

\begin{table*}[h]
    \fontsize{10pt}{10pt}
    \centering
    \begin{tabular}{c|c|c|c|c|c}
   \textbf{ML Algorithm} & \textbf{Feature Type} & \textbf{Trained Model} & \textbf{Article Split} & \textbf{Source Split} & \textbf{Country Split}\\
    \hline
       \multirow{6}{*}{\textbf{Random Forest}} & \multirow{3}{*}{\textbf{NELA}}  & \textbf{US vs. UR} & 0.904 (+/- 0.01) & 0.665 (+/- 0.21) & 0.488\\
        & & \textbf{UK vs. UR} & 0.873 (+/- 0.02) & 0.689 (+/- 0.12) & 0.784\\
        & & \textbf{US+UK vs. UR} & 0.850 (+/- 0.01) & 0.694 (+/- 0.28) & -\\
         \cline{2-6}
       & \multirow{3}{*}{\textbf{Doc2Vec}}  & \textbf{US vs. UR}  & 0.870 (+/- 0.02) & 0.635 (+/- 0.24) & 0.682 \\
        & & \textbf{UK vs. UR} & 0.874 (+/- 0.01) & 0.642 (+/- 0.22) & 0.612 \\
        & & \textbf{US+UK vs. UR} & 0.822 (+/- 0.01) & 0.815 (+/- 0.19) & -\\
        \hline
        \hline
        \multirow{6}{*}{\textbf{Random Forest}} & \multirow{3}{*}{\textbf{NELA-scaled}}  & \textbf{US vs. UR} & 0.860 (+/- 0.02) & 0.437 (+/- 0.37) & 0.755\\
        & & \textbf{UK vs. UR} & 0.854 (+/- 0.01) & 0.440 (+/- 0.28)  & 0.820\\
        & & \textbf{US+UK vs. UR} & 0.821 (+/- 0.01) & 0.570 (+/- 0.35) & -\\
         \cline{2-6}
       & \multirow{3}{*}{\textbf{Doc2Vec-scaled}}  & \textbf{US vs. UR}  & 0.850 (+/- 0.01) & 0.611 (+/- 0.27) & 0.742 \\
        & & \textbf{UK vs. UR} & 0.861 (+/- 0.01) & 0.610 (+/- 0.25) & 0.700\\
        & & \textbf{US+UK vs. UR} & 0.815 (+/- 0.01) & 0.689 (+/- 0.24) & -\\
        \hline
        \hline
        \multirow{6}{*}{\textbf{SVM}} &\multirow{3}{*}{\textbf{NELA-scaled}}  & \textbf{US vs. UR} & 0.850 (+/- 0.01) & 0.790 (+/- 0.14) & 0.413\\
         & & \textbf{UK vs. UR} & 0.850 (+/- 0.01) & 0.762 (+/- 0.12)  & 0.648\\
         & & \textbf{US+UK vs. UR} & 0.907 (+/- 0.00) & 0.689 (+/- 0.16) & -\\
         \cline{2-6}
       & \multirow{3}{*}{\textbf{Doc2Vec-scaled}}  & \textbf{US vs. UR}  & 0.885 (+/- 0.01) & 0.787 (+/- 0.11) & 0.621 \\
         & & \textbf{UK vs. UR} & 0.884 (+/- 0.00) & 0.857 (+/- 0.08) & 0.532\\
         & & \textbf{US+UK vs. UR} & 0.930 (+/- 0.00) & 0.804 (+/- 0.14) & -\\
    \end{tabular}
    \caption{Average accuracy and standard deviation over 20 folds per model per test scheme. The Country Split is the only test scheme that does not use folding. Each testing scheme is described in Section~\ref{sec:testing}. Note, NELA-scaled and Doc2Vec-scaled are both scaled and normalized.}
    \label{tab:results}
\end{table*}

\section{Results}
In Table \ref{tab:results}, we show accuracy results for each scheme and model. In Figure \ref{fig:conf}, we show the average, normalized confusion matrix for selected models under the source split scenario, showing which classes the models make mistakes on.

\textbf{Text-based news veracity models do not generalize across the US and the UK.} As we hypothesized, models trained on reliable US news do not perform well when tested on reliable UK news, and vice versa. Both when using Random Forest and SVM, the country split scenario results in accuracy less than random chance. When training a model on US reliable data and testing on UK reliable data we achieve only 48.8\% accuracy using Random Forest with NELA features and 41.3\% when using an SVM with NELA-scaled features, creating an accuracy drop of 41.6\% and 43.7\% from the article split scenario, respectively.

Interestingly, the low performance across countries is not consistent across all models or training sets. For example, when using Random Forest with scaled and normalized features (both for NELA and Doc2Vec), we see small accuracy drops between the article split and the country split. Specifically, when training a model using UK reliable data and testing on US reliable data, we achieve 82\% accuracy, only a drop of 3.4\% from the article split scenario.

We also see that the country split tests are not always symmetrical. Across all NELA models, the country split for models trained on US is much worse than those trained on UK. For example, when using Random Forest on NELA, we achieve 48.8\% accuracy when training on US and testing on UK, but we achieve a 78.4\% accuracy when training on UK and testing on US. When looking at the feature importance scores in the Random Forest model for each, we note that while the feature importance rankings are fairly similar, the feature importance scores are much more uniform across features in the UK trained model than in the US trained model. We suspect that this less skewed distribution of feature splits in the UK model may help generalization to the US, particularly since similar features are most important across both the models.  

\textbf{Text-based models have trouble generalizing to unseen, unreliable sources.}
In general, we see that all model combinations perform worse on unseen sources than on seen sources. The worst performing models are those using Random Forest with scaled and normalized features. Specifically, the best performing model only obtains 44.0\% accuracy when classifying unseen sources for NELA-scaled and 61.0\% accuracy for Doc2Vec-scaled, whereas the corresponding models with no scaling and normalization obtain 68.9\% accuracy and 64.2\% accuracy respectively. So, while scaling and normalizing features for Random Forest provides the best country split performance (as discussed above), it comes at a large cost in the source split test. 

This result seems counter-intuitive at first, as both test scenarios are classifying unseen sources. However, when looking at the confusion matrices in Figure~\ref{fig:conf}, we notice that the problem is not classifying unseen reliable sources (US or UK), but actually classifying unseen unreliable sources. In all models, we misclassify unseen UR sources more frequently, sometimes much more frequently, than we misclassify US or UK sources. In Figure~\ref{fig:conf}, we show a selection of the models, but the pattern remains consistent across all models in the source split scenario.

We partially attribute the trouble in classifying unseen, unreliable sources to the wide range in writing styles across these sources. When looking at the individual NELA feature distributions for each unreliable source (not shown in the paper), we see a much wider spread between sources than between the US or UK sources. This variation makes sense, as many unreliable sources do not have stylistic guidelines or copy editing services, in addition to employing individuals with little training in journalistic code of conduct. On the other hand, larger, mainstream sources are likely to have journalistic training and guidelines. 

\textbf{SVMs generalize to unseen, unreliable sources better than Random Forest.} Next, we note that the SVM models, both with NELA-scaled and with Doc2Vec-scaled, perform best across seen sources and unseen sources, with the smallest drop in accuracy between the article split test and the source split test being only 2.7\% when using Doc2Vec-scaled. Across the SVM models we also note a lower standard deviation over the folds, which may indicate SVM is over-fitting less to specific sources in the training set than Random Forest. Additionally, the confusion matrices for the SVM models support their higher performance, achieving less false positives than other classifiers. 


\textbf{Simply combining US and UK data in the training set is not good enough.} Lastly, we note that training with both US and UK data in the reliable class is not necessarily the solution to creating a better model. While the overall accuracy is not significantly harmed across all models settings, the number of false positives do increase. Some examples of this can be found in Figure~\ref{fig:conf}b, d, and f. We find this pattern consistent across all model settings.

\section{Conclusion and Future Work}
In conclusion, this short paper demonstrates that text-based news veracity models do not generalize to news media across different countries, even of the same language. Through a series of model testing scenarios, we show that not only do text-based models perform poorly across countries, but they can also perform poorly on unseen sources, particularly unseen unreliable sources. 

These results suggest several areas for future work. First, there is a need to understand what type of model are best for universal news veracity detection. Is it best to have separate models for each country? Can models be fine-tuned to specific countries? Questions such as these are open work. Second, these results point to the need to find features/representations that are, in a sense, orthogonal to text-based representations or news articles. It is clear that text-based models have some merit, but it is also clear they have many potential downfalls. If news veracity models begin to incorporate features based on signals outside of the text, such as source behavior or web page features, these models could be strengthened. 

There are limitations with this short study. First, due to the need to balance between the US and UK news data available in NELA-GT-2018, we use much less data than other models in the literature. In particular, the work in \cite{horne2019robust} shows much higher performance on unseen sources than we show in this work using similar models, but much more data. Hence, the models discussed in this paper can likely be improved with a larger and more diverse training data set, particularly for unreliable news sources. Second, we do not have a comparable unreliable US news and unreliable UK news. Of the unreliable news sources randomly selected, the majority of them do not have a known location, but of those that do (Infowars and True Pundit) are US based.

\small{\section{Acknowledgement}
This work was supported by the Rensselaer-IBM AI Research Collaboration (\url{http://airc.rpi.edu}), part of the IBM AI Horizons Network (\url{http://ibm.biz/AIHorizons}).
}
\bibliographystyle{aaai}
\bibliography{references}

\end{document}